\pdfoutput=1

\documentclass[11pt]{article}

\usepackage{acl}

\usepackage{times}
\usepackage{latexsym}
\usepackage{graphicx}
\usepackage{multirow}
\usepackage{bbm}
\usepackage[T1]{fontenc}
\usepackage{amsmath}
\usepackage[utf8]{inputenc}
\usepackage{booktabs}

\usepackage{microtype}

\title{How Interpretable are Reasoning Explanations \\from Prompting Large Language Models?}
\author{Wei Jie Yeo\textsuperscript{1}, Ranjan Satapathy\textsuperscript{2},  Rick Siow Mong Goh\textsuperscript{2}, Erik Cambria\textsuperscript{1}\\
\textsuperscript{1}School of Computer Science and Engineering, Nanyang Technological University, Singapore \\
\textsuperscript{2}Institute of High Performance Computing (IHPC), Agency for Science, Technology and Research (A$\textasteriskcentered$ STAR), \\1 Fusionopolis Way, \#16-16 Connexis, 138632, Singapore \\
\textsuperscript{1}\texttt{yeow0082@e.ntu.edu.sg},\\
\textsuperscript{2}\texttt{\{satapathy\_ranjan,gohsm\}@ihpc.a\-star.edu.sg},\\
\textsuperscript{1}\texttt{cambria@ntu.edu.sg}
}

\begin{document}
\maketitle
\begin{abstract}
Prompt Engineering has garnered significant attention for enhancing the performance of large language models across a multitude of tasks. Techniques such as the Chain-of-Thought not only bolster task performance but also delineate a clear trajectory of reasoning steps, offering a tangible form of explanation for the audience. Prior works on interpretability assess the reasoning chains yielded by Chain-of-Thought solely along a singular axis, namely faithfulness. We present a comprehensive and multifaceted evaluation of interpretability, examining not only faithfulness but also robustness and utility across multiple commonsense reasoning benchmarks. Likewise, our investigation is not confined to a single prompting technique; it expansively covers a multitude of prevalent prompting techniques employed in large language models, thereby ensuring a wide-ranging and exhaustive evaluation. In addition, we introduce a simple interpretability alignment technique, termed Self-Entailment-Alignment Chain-of-thought, that yields more than 70\% improvements across multiple dimensions of interpretability. Code is available at \url{https://github.com/SenticNet/CoT_interpretability}

\end{abstract}

\section{Introduction}

In recent trends, Large Language Models (LLM) have shown impressive performance across a diverse array of tasks, primarily through extensive scaling of model size~\cite{brown2020language}. Techniques such as instruct-tuning~\cite{wei2021finetuned} applied across diverse tasks have empowered LLMs to execute inference on previously unseen tasks. One attributing factor lies with the extensive efforts put into innovating new ways of prompting the LLM to better exploit their knowledge base.
Chain-of-Thought (CoT)~\cite{wei2022chain} has gathered much attention due to its simple setup which allows the LLM to generate not only the task output but also the steps undertaken. 

In addition to its efficacy in enhancing the model’s performance, this prompting method concurrently touches on one of the important aspects of utilizing these models for decision-making: interpretability. 
The assumption is that the reasoning chain preceding the answer illustrates the model's thought process, enabling the audience to understand how the answer is derived. However, such claims though seemingly plausible should be taken lightly as they may not be faithful to the model's reasoning process~\cite{jacovi2020towards}. In this context, \textit{plausibility} refers to the extent to which an explanation resonates with and is deemed acceptable by a human audience. \textit{Faithfulness}, on the other hand, is characterized by the extent to which the explanation accurately reflects the model's decision-making process.

There has been a large number of works that seek to introduce modifications to CoT, including Self-Consistency~\cite{wang2022self}, Least-to-Most~\cite{zhou2022least}, while others specifically focus on establishing faithful reasoning~\cite{creswell2022faithful,lyu2023faithful}. We introduce a simple extension to the list of CoT variants, but purely with a focus on enhancing interpretability in the reasoning chain. The approach coined \textit{Self-Entailment-Alignment CoT (SEA-CoT)} operates similarly to Self-Consistency, but additionally utilizes a form of consistency between the corresponding reasoning steps and supported context. This action is missing in Self-Consistency, as the focus is only on the resultant output, potentially leading to unfaithful reasoning which may not support the underlying answer.  

Moreover, we conduct an extensive investigation into the reasoning explanations by evaluating under three pivotal axes of interpretability: faithfulness, robustness, and utility on three commonsense reasoning datasets. These assessments are implemented across multiple prompting techniques including CoT and various adaptations of it. 

\section{Motivation}
\label{sec:2}
Efforts aimed to enhance faithfulness in NLP take various forms. Extractive rationalizing model~\cite{lei2016rationalizing}, designed to be faithful, generally comprises two separate components: explainer and predictor. This design paradigm conditions the predictor exclusively on text spans extracted by the explainer, positing that the resultant output, $\hat{y}$ is faithfully aligned with the extracted text, $\hat{e}$. However, prior studies~\cite{wiegreffe2020measuring} cautions against such beliefs, identifying limitations in adopting the explain-then-predict approach. The authors mentioned that such an approach restricts the focus of the predictor toward the target identified by the explainer, thereby raising questions about what is being explained. Conversely, Jacovi et al.~\cite{jacovi2021aligning} highlight concerns relating to the lack of meaningful insights from multiple text spans.

In accordance, we hypothesize that besides the limitation of narrowing the predictors' context, generating the explanation and output using separate modules could compromise the quality of the explanation. We set up a simple study, comparing a modular against a single LLM setup on two interpretability traits, faithfulness, and utility, covered in deeper detail in section \ref{sec:interpretability}. We adopt the PINTO framework~\cite{wang2022pinto}, where the explainer, $r_{\theta}$ is a frozen pre-trained LLM and the predictor, $f_{\phi}$ is finetuned on the downstream task, conditioned on both the generated explanation and context, $\hat{y} = f_{\phi}(x \oplus \hat{e})$, where $\hat{e} = r_{\theta}(x)$, x is the given context and $\oplus$ is the concatenation process.

For the single LLM setup, we directly train $f_{\phi}$ to generate both $\hat{e}$ and $\hat{y}$ jointly. We measure faithfulness by computing the drop in performance when swapping $\hat{e_i}$ with another instance within the same batch, $\hat{e_{j \neq i}}$ before deriving $\hat{y}|x;\hat{e}$. We use Leakage-Adjusted Simulatability (LAS)~\cite{hase2020leakage}, to measure the utility of the rationale, a higher score would indicate that $\hat{e}$ is more useful towards learning $\hat{y}$. The details of LAS are covered in ~\ref{appendix:LAS}
\begin{figure}[h]
    \centering 
    \includegraphics[width=0.5\textwidth]{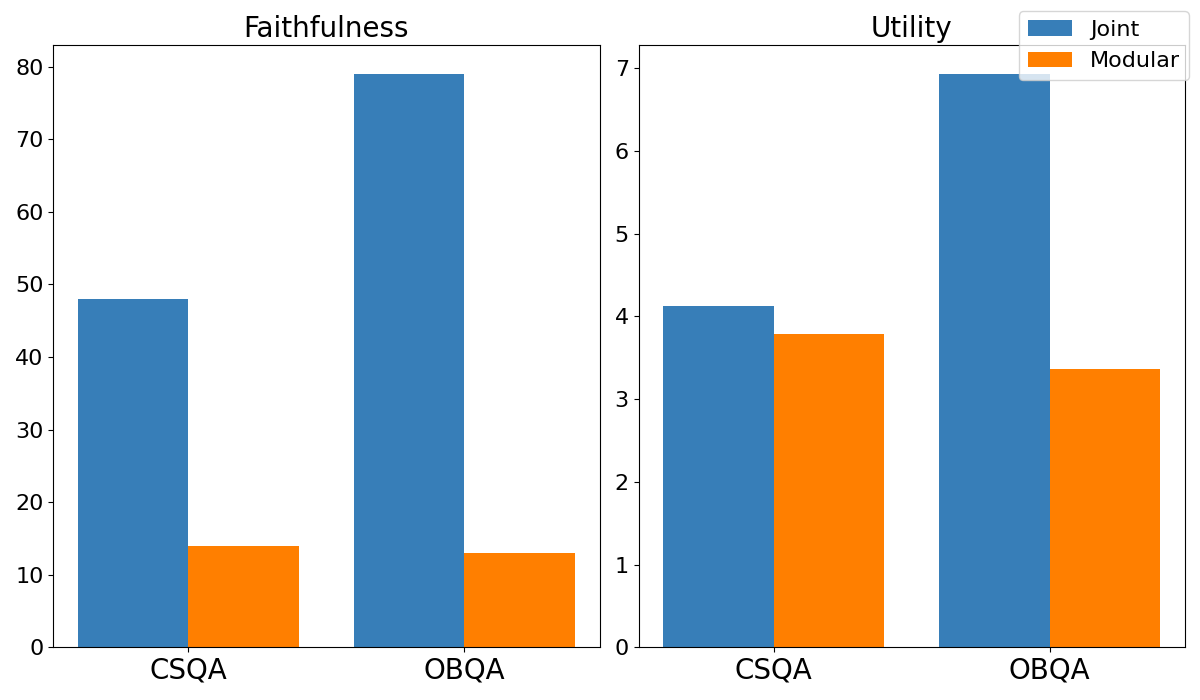}
    \caption{Faithfulness and Utility scores for joint and modular approach on two reasoning datasets: CSQA and OBQA.} 
    \label{fig:prelim}
\end{figure}

We conduct experiments on two commonsense reasoning datasets: Commonsense QA (CSQA)~\cite{talmor2018commonsenseqa} and OpenBookQA (OBQA)~\cite{mihaylov2018can}.
Figure~\ref{fig:prelim} shows that the joint approach scores higher on both accounts of faithfulness and utility. We hypothesize that a single model is in better control of aligning its explanation to the resultant outcome. Contrarily, a model relying on explanations synthesized by an external model may instead exhibit a diminished correlation between the interdependent variables, explaining the marginal difference in performance despite being given an unrelated stimulus.

Notably, this observation resonates well with the recognized capability of recent LLMs to self-generate text serving diverse objectives. In particular, LLMs pre-trained on a large amount of text can elucidate their reasoning processes, assisted with the appropriate prompting format. This preliminary experiment serves as the main motivation to conduct experiments to scrutinize the quality of explanations produced by a singular LLM.

\section{Prompt Techniques}
\label{sec:CoT}
\begin{figure*}[ht]
    \centering 
    \includegraphics[width=0.9\textwidth]{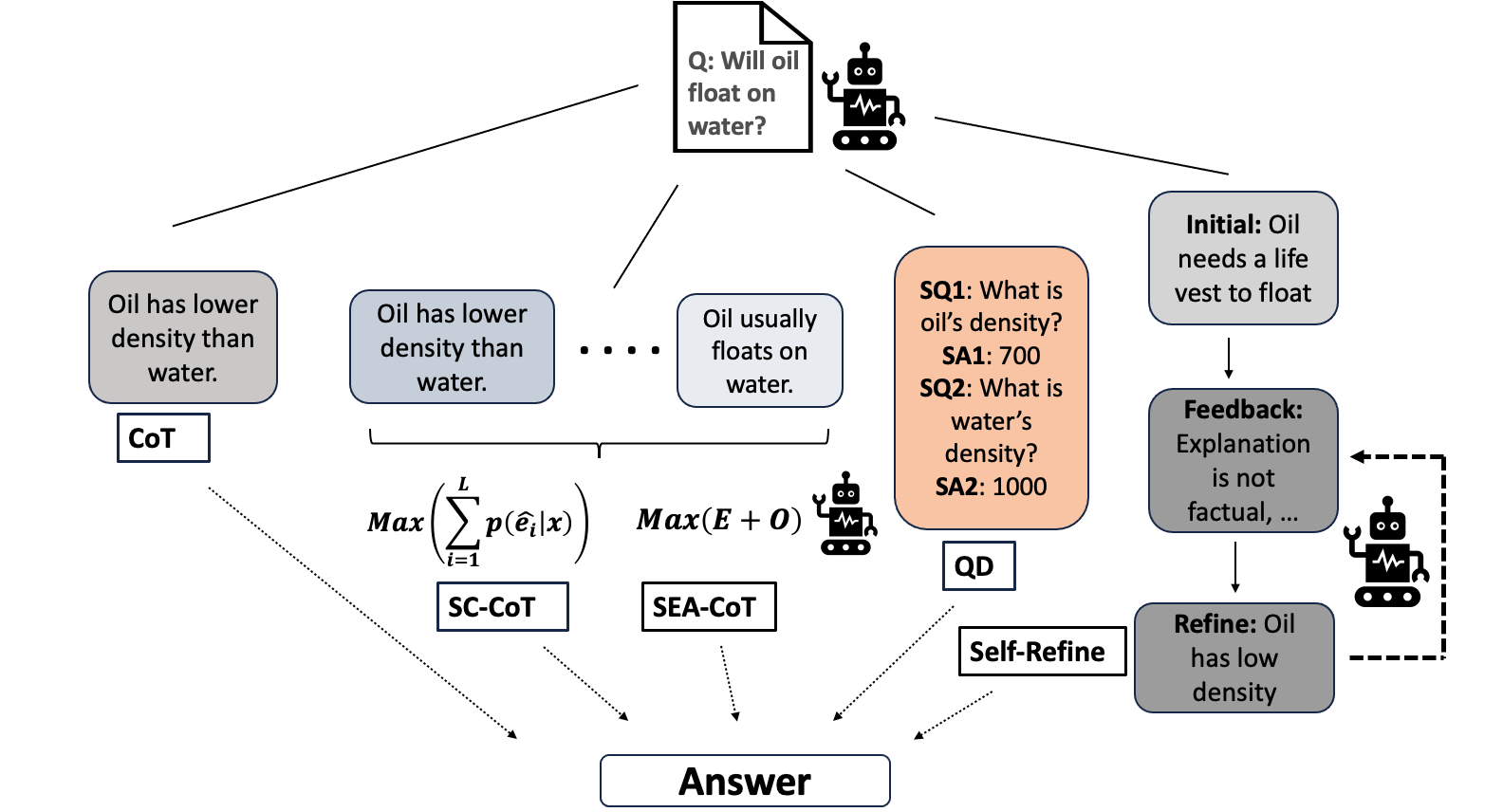}
    \caption{Overview of different prompting techniques to derive the reasoning chain, to serve as the explanation (boxed with dashed line). [Left to Right]: Cot, SC-CoT, SEA-CoT, QD, and Self-Refine (SR). SC-CoT and SEC-CoT differ in the explanation selection stage, where the former selects based on maximum cumulative probability and the latter on two objectives: entailment, $E$, and overlap, $O$ with an additional forward pass. Each robot figure denotes a forward pass from the LLM, SR stops when a stopping criteria is encountered or exceeds the max number of passes. SR requires the most pass, 3 per round.} 
    \label{fig:overview}
\end{figure*}
In this section, we systematically review various ways an LLM, $f_{\phi}$ can be prompted. These methods primarily differ in how the language model is queried to derive the final answer. Furthermore, we proposed an approach, SEA-CoT, aimed at improving the interpretability traits of the reasoning chain to serve as the explanation for the resultant output. A high-level overview is shown in Figure~\ref{fig:overview}.
\begin{itemize}
    \item \textbf{CoT}: Chain-of-thought prompting has shown promising results in encouraging an LLM to better answer the task by reasoning aloud the steps before arriving at the final answer. ~\cite{kojima2022large} has shown that it is possible in the zero-shot setting simply by appending \textit{`Let's think step by step'} at the end of the instruction. 
    
    \item \textbf{Self-Consistent CoT (SC-CoT)}: Following on, other works like Self-Consistency~\cite{wang2022self} address the suboptimality of greedy decoding in CoT by sampling multiple, $N$ paths and choosing the final answer, $\hat{y}^*$ via majority voting. SC-CoT has shown improvements across multiple arithmetics and commonsense reasoning benchmarks. Since multiple explanations may lead to the majority answer. We choose the explanation with the highest cumulative probability. We also experiment with different ways, further discussed in the ablation section.
    
    \item \textbf{Question decomposition (QD)}: ~\cite{zhou2022least} demonstrates that decomposing a complex problem into more manageable sub-problems significantly facilitates the problem-solving capability of the model. The model answers each sub-problem and pieces together the answers to conclude the principal problem. We treat the sub-question and answers as the target explanation and assess their interpretability properties.
    
    \item \textbf{Self-Refine (SR)}: SR~\cite{madaan2023self} is a type of iterative process of prompting the LLM with a set of instructions. The main idea is to instruct the LLM to continuously provide feedback for its' own output and refine using the feedback, the process stops when the feedback deems the output as sufficient in solving the task at hand. The whole iterative process is achieved by self-prompting the same language model. There exist other forms of acquiring feedback, such as querying a trained feedback model or using external factual knowledge~\cite{pan2023automatically}. We choose the approach of querying the same LLM as we are focused on the explainability of generated outputs from a sole LLM. 
\end{itemize}

\section{Proposed Approach}
Most adaptations on CoT are only aimed at maximizing task performance as covered in Section \ref{sec:CoT}. Our work is instead focused on enhancing the interpretability of the presented reasoning chain preceding the task output. We adapt from SC-CoT, by focusing on the $N$ sequences produced, ranked based on specified objectives. Instead of picking explanations based on heuristics such as highest cumulative probability, the reasoning is chosen based on the maximization of two objectives: entailment and overlapping score between the supported context $(x \oplus \hat{y})$ and reasoning $\hat{e}$. 

We posit that a credible explanation should intrinsically align with the given context it aims to elucidate~\cite{jie2024plausible}; in this scenario, it encompasses both the question being addressed and the prediction label, measured by the level of entailment. 

We additionally maximize the overlap between two sets of key tokens\footnote{The two sets of tokens are compared after removing any stopwords to minimize noise within the context}, which we show in later experiments to be beneficial towards producing higher quality explanations. This simple approach can be regarded as performing a self-alignment step to pick the most suitable explanation with the $N$ sequences.

Inspired by works that employ the LLM itself to do self-correction, we do the same by asking the LLM to rate the entailment level between each own generated reasoning, $\hat{e}_i$ and the joint context, $x \oplus \hat{y}$. We prompt the LLM with few-shot examples of natural language inference (NLI), $x_{e}$ in Figure~\ref{fig:entailment} and determine if the hypothesis entails the premise. The final score to be ranked, $S_T$ is a combination of both the probability of entailment, $S_e$, and the IoU score, $S_o$.
\begin{align}
    S_e & = p_e(f_{\phi}(x \oplus \hat{y}, \hat{e_i}|x_{e})) \\
    S_o & = \frac{|\hat{e_i} \cap (x \oplus \hat{y})|}{|\hat{e_i} \cup (x \oplus \hat{y})|} \\
    S_T & = S_e + S_o 
\end{align}
The most interpretable explanation is then chosen via maximizing $S_T$. One caveat is that in the event where $|\hat{y}^*| = 1$, we fall back to SC-CoT. However, this can be avoided by trivially setting the number of sequences, $N$ to be higher than the number of possible options.\\

\section{Interpretability Qualities}
\label{sec:interpretability}
\begin{figure}[h]
    \centering 
    \includegraphics[width=\columnwidth]{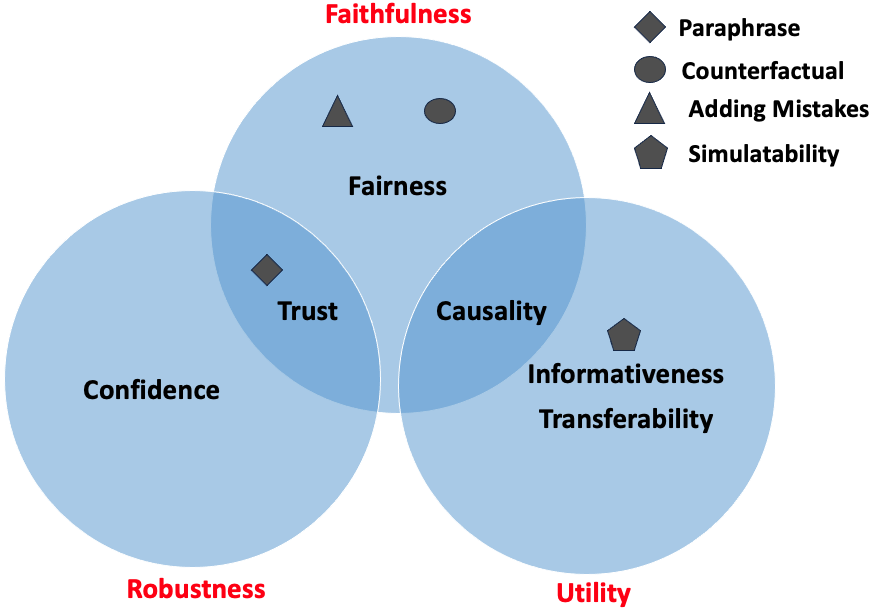}
    \caption{The interpretability qualities measured by different perturbation tests, to achieve the corresponding goals of an explanation. Goals referenced from~\cite{yeo2023comprehensive}}
    \label{fig:inter_quality}
\end{figure}

\begin{figure*}[ht]
    \centering 
    \includegraphics[width=0.82\textwidth]{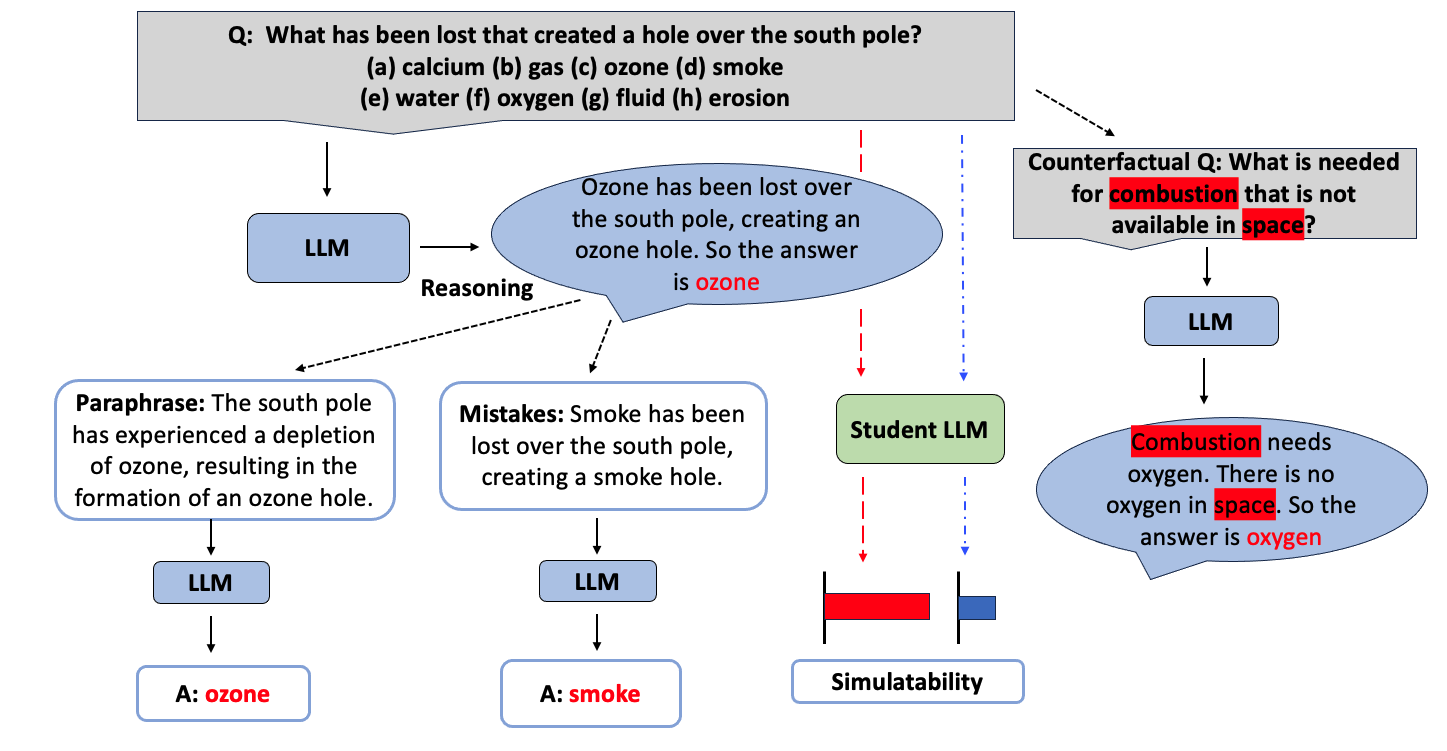}
    \caption{Interpretability test for faithfulness, robustness, and utility. The reasoning chain is subjected to perturbations: paraphrasing and inserting mistakes, before re-generating the subsequent output. Counterfactual: the original question is changed to check if the resultant reasoning accounts for edits (shaded red). Simulatability: increase in task performance when training data is augmented with reasoning chain, measured with a student model.} 
    \label{fig:permutation}
\end{figure*}

Interpretability is a multifaceted characteristic with multiple desirable traits concerning various goals of interpretability. Inspired by existing work on desirable goals of explainable AI~\cite{yeo2023comprehensive}, we assess three aspects of interpretability: faithfulness, robustness, and utility. We propose these traits as we believe they are directly linked to achieving such goals, illustrated in Figure~\ref{fig:inter_quality}. We discuss the connections in further sections. In accordance, we outline the corresponding evaluations sought out to assess each trait, shown in Figure ~\ref{fig:permutation}. These evaluations are primarily conditioned on both the context and self-generated reasoning chain.\\\\
\textbf{Faithfulness}: The concept of faithfulness seeks to gauge the extent to which the explanation aligns with the underlying decision-making process.~\cite{lanham2023measuring} conducted a series of tests assessing the faithfulness of reasoning chains generated using CoT from an LLM. However, the authors only investigated a single prompting technique, while we conducted extensive experiments covering multiple prompting approaches. A faithful explanation is crucial as it fosters trust~\cite{cambria2023seven} and fairness, ensuring that users can rely on the explanation to reflect the decision-making process and identify any potential biases, thereby improving model transparency and understanding of any causal relationships\\\\
\textbf{Robustness}: Robustness seeks to measure how resilient or consistent a given explanation is under various circumstances. For instance, employing adversarial attacks on an explanation, as delineated by ~\cite{chen2022can}, could serve as a mechanism to ascertain whether the model's decision is susceptible to diversion or distraction induced by these attacks. A robust explanation instills confidence and trust in users that the model would behave appropriately despite noises in the input.\\\\
\textbf{Utility}: A largely understudied but important trait, utility is paramount to maximizing the information conveyed to the audience. A useful explanation can allow the discovery of new knowledge to human users such as understanding the causal relationships or enable more efficient knowledge distillation between neural models.

\subsection{Paraphrase}
Paraphrasing $\hat{e}$ corresponding to $\hat{y}$ allows us to question the robustness of the explanation, ie how robust is the explanation against minor variations, assuming that these variations do not alter the core intent, yet still enable the model to produce the same outcome? Albeit such a test concurrently touches on the concept of faithfulness, where similar thought processes should lead to identical conclusions given the same model~\cite{jacovi2020towards}. However, for the sake of differentiation, we consider the primary objective of paraphrasing as an evaluation of robustness in the following experiments. 

\subsection{Adding mistakes}
\label{sec:adding mistakes}
In contrast to ensuring answer consistency among similar reasoning, inserting erroneous inputs into an explanation can assess if the reasoning preceding the output is truly faithful. One would expect the model to change its decision given an erroneous reasoning chain if it is faithful from the start. We focus on the alteration in prediction rather than actual task performance, since incorrect reasoning may potentially correct an erroneous explanation, though such occurrences are exceedingly rare. 

\subsection{Simulatability}
As it is costly to employ humans to assess if a reasoning chain is useful, we employ forward simulatability as a proxy for utility. We measure simulatability using LAS in Section~\ref{sec:2} (further details in ~\ref{appendix:LAS}) as it highly correlates with human judgment. A 220M T5-base~\cite{raffel2020exploring} is selected as the student model, to measure utility from improvement in downstream performance. The generated reasoning, $\hat{e}$ is appended to the input context $x$, which is then used as the final context for predicting the task label, $\hat{y} = f_s(\hat{e} \oplus x)$, where $f_s$ refers to the student model. The student model undergoes fine-tuning with the aid of these explanations, followed by an evaluation of its performance. A key aspect of LAS lies with the notion of subtracting a baseline, $M_s(f_s(x))$ from $M_s(f_s(\hat{e} \oplus x))$, where $M_s$ is a task scoring function such as accuracy or F1-score. This is used to assess the benefits gained by adding $\hat{e}$ into the training process.

\subsection{Counterfactual reasoning}
An alternative method to ascertain faithfulness follows by evaluating whether an explanation would change when the original question is modified in a different direction, particularly when directed towards a counterfactual scenario. \cite{atanasova2023faithfulness} shows that an instance of unfaithfulness can be detected if the counterfactual explanation, $e'$ does not acknowledge the modifications, $c$ in the counterfactual instance $x_i':y'$, yet still successfully predicting the counterfactual label, $y' \neq y$. The distinction from Section~\ref{sec:adding mistakes} is that besides detecting signs of unfaithfulness, it also embodies a directed approach that assesses a model's capacity to contemplate alternative scenarios. 

Conversely, introducing mistakes can be seen as an undirected measure aimed at gauging the decline in confidence, given an erroneous prior belief. We deemed an instance of unfaithfulness under the following conditions:
\begin{enumerate}
    \item $x_i' = \{x_{i,1},x_{i,2}...c,...x_{i,L}\}:y_i'$
    \item $\hat{y} = y  \land  \hat{y'} = y'$
    \item $e' \cap c = \emptyset$
\end{enumerate}
The first two conditions are prerequisites for assessment, while the third indicates signs of unfaithfulness.

\section{Experiments}
\textbf{Datasets}: We implement perturbation experiments across three commonsense reasoning benchmarks.
\begin{enumerate}
    \item OpenBookQA~\cite{mihaylov2018can}, which has 4 answer choices for each question and assesses open-book reasoning capabilities.
    \item QASC~\cite{khot2020qasc}, is an 8-choice multi-hop reasoning dataset, requiring assembling multiple real-world facts to successfully answer the question.  
    \item StrategyQA~\cite{geva2021did} is a binary question dataset structured such that the model is required to strategize a chain of reasoning steps to derive the correct answer. 
\end{enumerate}
We use only the test set to run the experiments for all perturbations introduced in Section~\ref{sec:interpretability}, except LAS, where we employ the LLM to generate explanations for the training set as well.\\\\
\textbf{Implementation details}: We use the 70B Llama-v2~\cite{touvron2023llama} from Meta as the choice of LLM for this experiment. We use a 4-bit quantized version, via applying the GPTQ technique~\cite{frantar2022gptq} since the full 32-bit model would require extensive resources. The full model implementation details can be found in Appendix~\ref{appendix: inference}.\\\\
\textbf{Explanation modifications}: We perform automatic checks on the modifications to prevent errors in the experiments. In paraphrase, the modified explanation is chosen only when the resultant output $\hat{y}|e_m$ remains the same, and the opposite for mistakes insertion. In counterfactual generation. OpenAI's GPT3.5 is used for both paraphrasing and mistake insertion and GPT4 for counterfactuals since the task is much harder as $x'$ has to correspond to an alternate answer choice.\\\\
\textbf{Metric details}: We use the percentage of flipped predictions as the measurement unit for both paraphrased and mistake insertion. For counterfactual inputs, we only consider $e$ to be unfaithful if the counterfactual part, $e'$ has a zero overlap with modification $c$. This applies to singular reasoning chains, except QD where we only assessed each sub-answer. Utility is measured using the LAS score, corresponding to the increase in performance when supplemented with explanation during training. We list the prompt templates in Appendix~\ref{appendix: perturbation prompts}. We compute an aggregate score, averaging across the four qualities, after normalizing each score between 0 and 1.  For paraphrase and counterfactual, we take the complement score, $1-s$, where $s$ is the original unit.

\subsection{Results}
\label{sec:results}
\begin{figure*}[ht]
    \centering 
    \includegraphics[width=0.9\textwidth]{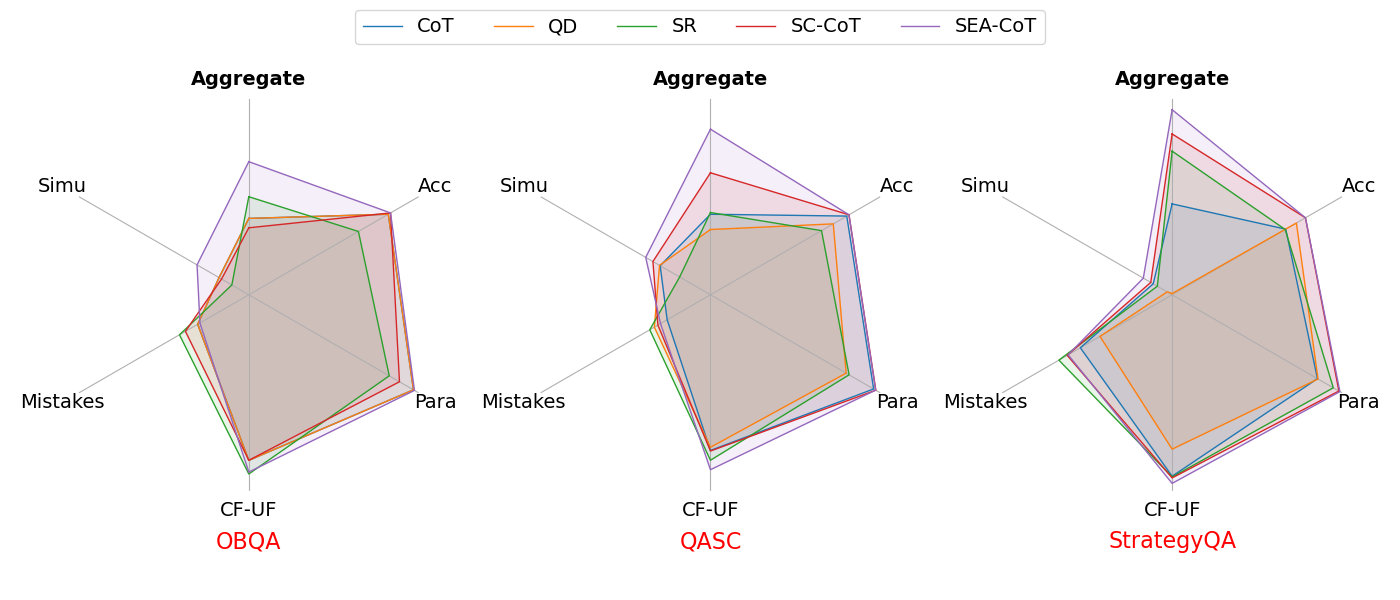}
    \caption{Interpretability results for the 5 prompting techniques across 3 commonsense reasoning benchmarks. Three axes of interpretability were assessed. 1) Robustness measured via paraphrasing (Para). 2) Faithfulness is measured with both counterfactual explanations (CF-UF) and mistake insertion. 3) Utility is represented using simulatability (Simu) of explanation. \textbf{Aggregate} is the combined average score across the three axes. CF-UF measures unfaithfulness instead of faithfulness. We take the complement of Para and CF-UF since a lower score is better.} 
    \label{fig:main_result}
\end{figure*}
We show the full experimental results in Figure~\ref{fig:main_result}. SEA-CoT surpasses all other baseline methods based on the average normalized score, notably displaying a significant difference in OBQA ($> 75 \%$) over majority of the baselines. Although SC-CoT is competitive, it still underperforms substantially as compared to SEA-CoT. We observe that the underperformance of SEA-CoT in the mistakes criteria can be explained via the relationship between SR's weak task performance and high score in mistakes, attributed to a higher likelihood of altering its output. Whereas, SEA-CoT achieves the highest task performance, albeit causing a trade-off in this regard. Nonetheless, despite comparable levels of task performance, SEA-CoT consistently surpasses SC-CoT across other metrics, indicating that the superior score achieved is still dependent on the selected reasoning.

\begin{figure}[ht]
    \centering 
    \includegraphics[width=0.5\textwidth]{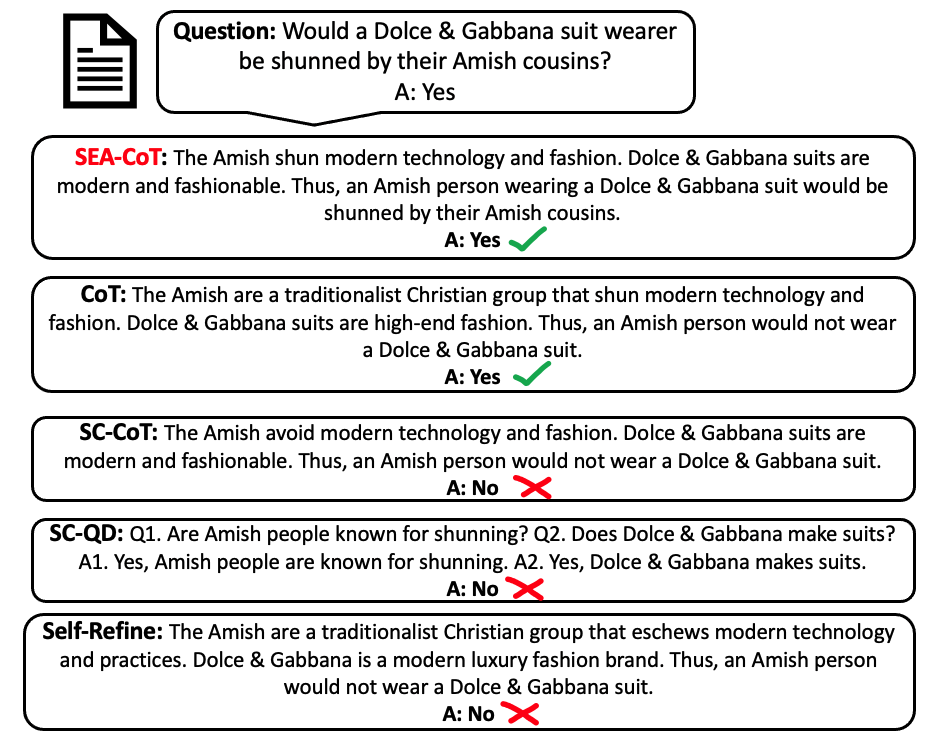}
    \caption{StrategyQA example, the reasoning chain produced by SEA-CoT reflects the important points in the context, making it easier for a learner model to simulate the answer from the given explanation.}
    \label{fig:utility}
\end{figure}
The key distinction between SC-CoT and SEA-CoT is the latter's self-critique step, which evaluates how its explanations align with the context and the intended answer. This approach significantly boosts utility and reduces unfaithfulness in counterfactual contexts. Higher utility scores support the hypothesis that context-aligned stimuli enhance the efficiency of learning signals, facilitating easier training for student models. 

Looking closer in Figure~\ref{fig:utility}, where the word \textit{"shunned"} is mentioned while other baselines used \textit{"would not wear"}, which does not directly relate to the target question, causing the model to erroneously infer the wrong label. While CoT successfully determines the correct answer, it fails to acknowledge the mention of \textit{"Amish cousins"}, thus exhibiting a tenuous connection to the question.
Unexpectedly, Self-Refine underperforms compared to other baselines, aligning with~\cite{huang2023large} who highlight the drawbacks of self-correction in reasoning tasks. 

The primary challenge stems from the intricacy of designing few-shot examples that can effectively drive successive enhancements over prior outputs, limiting the potential for self-improvement. 
SEA-CoT, however, not only prompts self-assessment but also offers targeted guidance to enhance reasoning consistency with the relevant context. This simple extension greatly improves the quality of the explanation, with no downside on performance.

\subsection{Ablation}
\label{sec: ablation}
\begin{table}[ht]
\centering
\small
\begin{tabular}{ccccc}
\toprule
Type & P($\downarrow$) & CF-UF ($\downarrow$) & M ($\uparrow$) & S ($\uparrow$)\\
\midrule
Random & 6.1 & 6.44 & 62.17 & 11.87 \\
Max & 1.8 & 6.6 & 61.8 & 12.59 \\
Overlap (O) & 1.56 & 5.04 & 70.83 & 14.88 \\
Entailment (E) & 2.38 & 5.46 & 69.99 & 13.46 \\
O\&E (SEA-CoT) & \textbf{1.2} & \textbf{3.81} & 61.24 & \textbf{16.97} \\
\bottomrule
\end{tabular}
\caption{Ablation over ways of selecting reasoning steps to serve as an explanation on StrategyQA. (O\&E) is the proposed SEA-CoT which uses both components.}
\label{tab:ablation}
\end{table}

This ablation seeks to study the effectiveness of choosing the most suitable reasoning chain. We break down SEA-CoT's ranking components and assess each of them, namely the entailment and keyword overlapping score. We additionally implement a baseline of SC-CoT that randomly picks from the list of explanations corresponding to the majority answer. The results from Table~\ref{tab:ablation} demonstrate the efficacy of considering both components of SEA-CoT when ranking reasoning explanations. 

Choosing the most probable reasoning step has shown to not perform well, whereas our approach targeted at enhancing the important traits of an explanation is simple and yet does not hinder performance. We also conduct additional studies on different values of N sequences in Table~\ref{tab:N values}.
\subsection{Model size}
\begin{table}[ht]
\centering
\small
\begin{tabular}{ccccc}
\toprule
Size & P($\downarrow$) & CF-UF ($\downarrow$) & M ($\uparrow$) & S ($\uparrow$)\\
\midrule
70B & \textbf{1.2} & \textbf{3.81} & 61.24 & \textbf{16.97} \\
13B & 4.1 & 4.38 & 69.62 & 6.16 \\
7B & 3.79 & 7.81 & \textbf{70.62} & 15.97 \\
\bottomrule
\end{tabular}
\caption{Interpretability scores between different model sizes}
\label{tab:model sizes}
\end{table}The scaling laws of model size primarily concern the downstream performance of LLMs but little is known regarding the influence on interpretability properties. We replicate the experiments on the StrategyQA dataset with a focus on SEA-CoT prompting.
We present the results in Table~\ref{tab:model sizes}. The largest model, 70B generally outperforms the smaller sizes across all metrics while observing the same phenomenon in mistake insertion, previously discussed in \ref{sec:results}. The improvement over smaller sizes may also be attributed to the enhanced accuracy in generating entailment scores for the explanation, analogous to observing greater performance of larger models in NLI tasks. Llama-13B surprisingly performs worse than its smaller variant, despite having a bigger network. Moreover, we note that by using SEA-CoT, even a 7B-sized model can generate more interpretable reasoning chains than a 70B model with other baseline prompts.

\section{Related Works}
\textbf{Natural Language Explanation (NLE)}: NLE can primarily be categorized as either abstractive (AE) or extractive (EE). The former is unrestricted by the context and as such produces more plausible explanations, while the latter is aimed at ensuring faithfulness. Notably, EE falls short in the realm of plausibility since humans do not understand spans of text without a full context in view~\cite{gurrapu2023rationalization}.
\cite{majumder2021knowledge} utilizes a union of both forms, conditioning the generation of AE on the extracted spans of text while concurrently grounding the generation on relevant world knowledge. Similar works include faithfulness through task decomposition~\cite{sanyal2022fairr}, label-specific explanations~\cite{kumar2020nile}. \cite{narang2020wt5} demonstrate the possibility of zero-shot explanation generation by pretending the word \textit{explain} to the input prompt.\\\\
\textbf{Interpretable CoT}: Since its introduction, CoT has garnered interest in the research community to innovate adaptation of it~\cite{chu2023survey}. Despite CoT being primarily introduced to improve reasoning skills of LLMs, there is much interest to see if these reasoning steps could be used to explain the model's thought process. Most works primarily investigate faithfulness of the reasoning~\cite{lanham2023measuring,radhakrishnan2023question,turpin2023language} or improving the faithfulness in CoT outputs, via refinement through knowledge retrieval~\cite{he2022rethinking}, symbolic reasoning~\cite{lyu2023faithful}, iterative information selection~\cite{creswell2022faithful} and factuality calibration~\cite{ye2022unreliability}. 
Other works focus on ascertaining the faithfulness of an explanation to the presence of factuality~\cite{wang2023knowledge,he2022rethinking,prasad2023receval}. While factuality is an important trait, it is not a sufficient component to ascertain faithfulness. Non-factual explanations may still align faithfully with an incorrect answer. Other works concentrate on semantic correctness~\cite{golovneva2022roscoe}, regarded closer to plausibility, which differs from the traits assessed in this study. Our work strives to conduct a holistic assessment of interpretability across various forms of prompting techniques used in LLMs, taking into account multiple properties that may be of importance to various audiences.

\section{Conclusion}
This work studied multiple ways of assessing the interpretability of an explanation. Our work is centered on assessing different variants of CoT and how we can better determine the suitability of the reasoning by-product as an explanation for the underlying prediction. We also propose a modification to the SC-CoT framework called \textit{SEA-CoT}, designed specifically to yield explanations that better fulfill the objectives of interpretability.  Our proposed framework surpasses the Robustness, Faithfulness, and Utility dimensions across multiple reasoning benchmarks. In the future, we plan to extend our work towards instilling interpretability and safety in the training stages~\cite{yang2023harnessing}, such as safety alignment in LLM. 

\section{Limitations}
Our work only investigates a single LLM - Llama-2 This work could be extended toward transformers of different structures such as encoder or encoder-decoder, or larger models, such as GPT3.5/4.0, which due to limiting resources are restricted to generate modifications instead. A secondary limitation is the quality of modifications to the original explanation, though we ask the modifier to check the outcome of the modified inputs (i.e. output remains the same when paraphrased), the correctness is nonetheless subjected to the ability of the modifier. This work did not study techniques that ground the LLM's response via external knowledge, which we note is an interesting avenue to consider next. An inherent weakness in LLM is self hallucination where it produces plausible text which are non-factual. Our work also left out investigating hybrid approaches such as Neuro-symbolic AI, which combines the learning abilities of neural networks and inherent interpretable decision-making frameworks of symbolic AI.

\bibliography{anthology,custom}

\appendix

\section{Appendix}
\label{sec:appendix}
\subsection{Perturbation details}
\label{appendix: perturbation prompts}
\begin{figure*}[h]
    \centering 
    \includegraphics[width=1.0\textwidth]{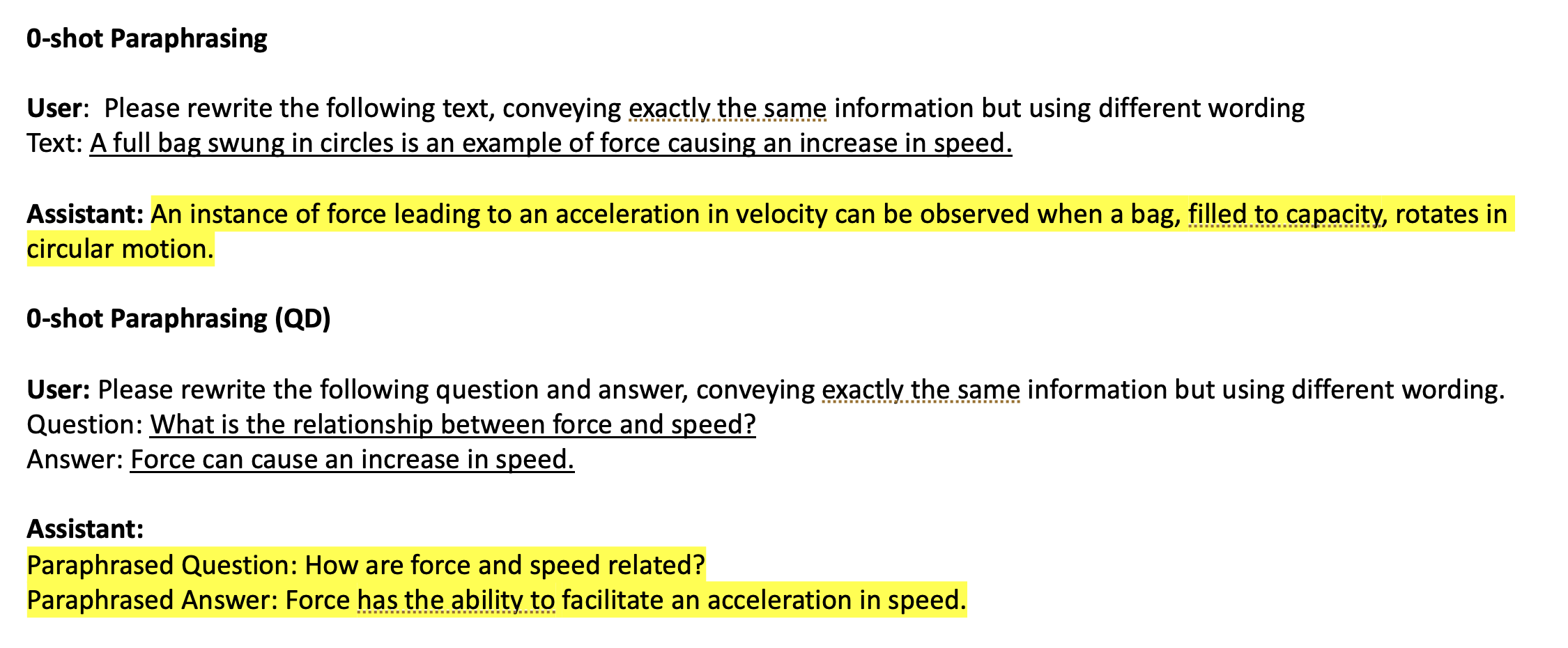}
    \caption{0-shot paraphrase template. Input [Underline] Generated: [highlighted]} 
    \label{fig:para}
\end{figure*}

\begin{figure*}[h]
    \centering 
    \includegraphics[width=1.0\textwidth]{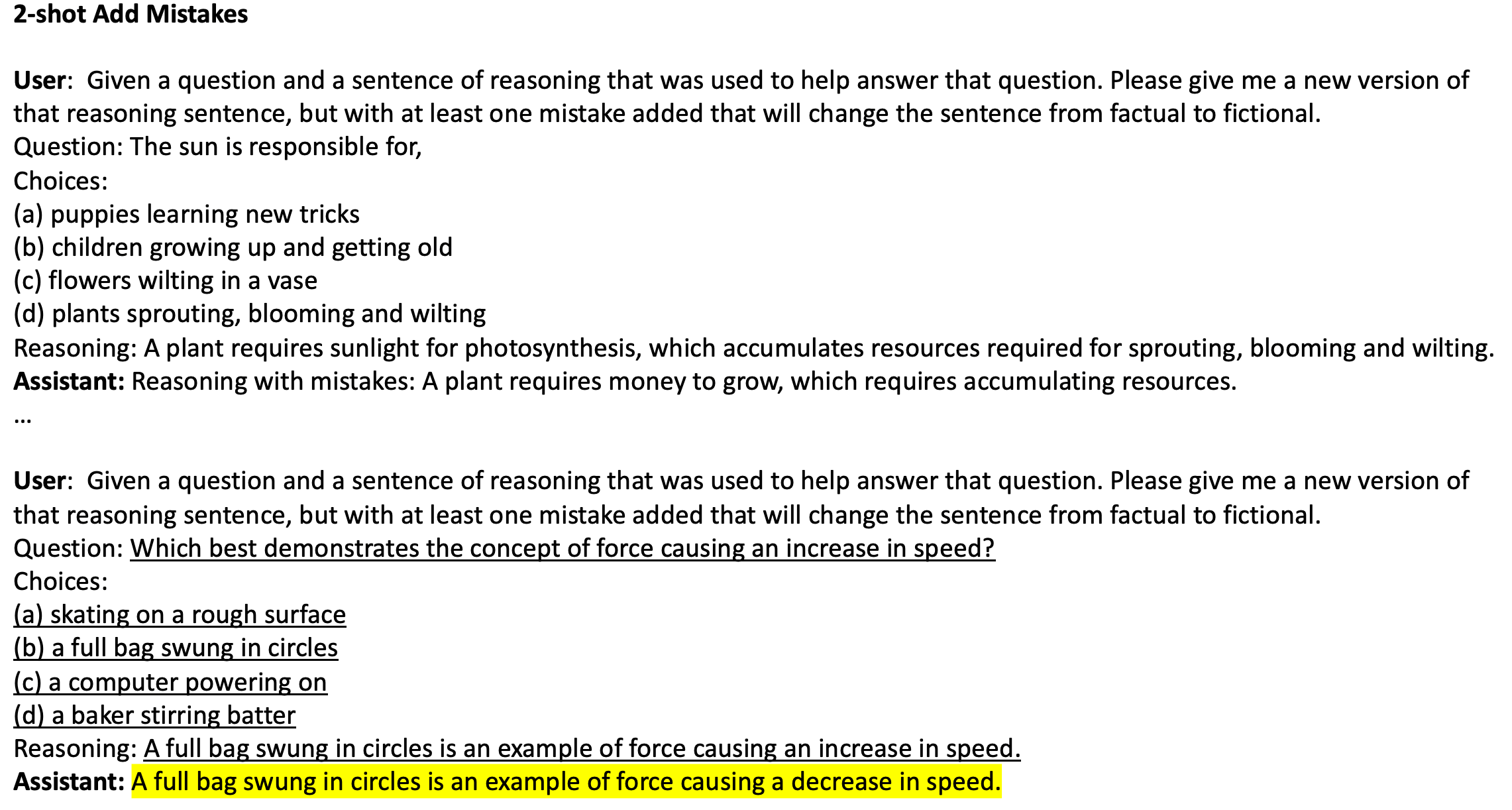}
    \caption{2-shot inserting mistake template for all prompts except QD. Input [Underline] Generated: [highlighted]. Only show 1 example.} 
    \label{fig:mistake}
\end{figure*}

\begin{figure*}[h]
    \centering 
    \includegraphics[width=1.0\textwidth]{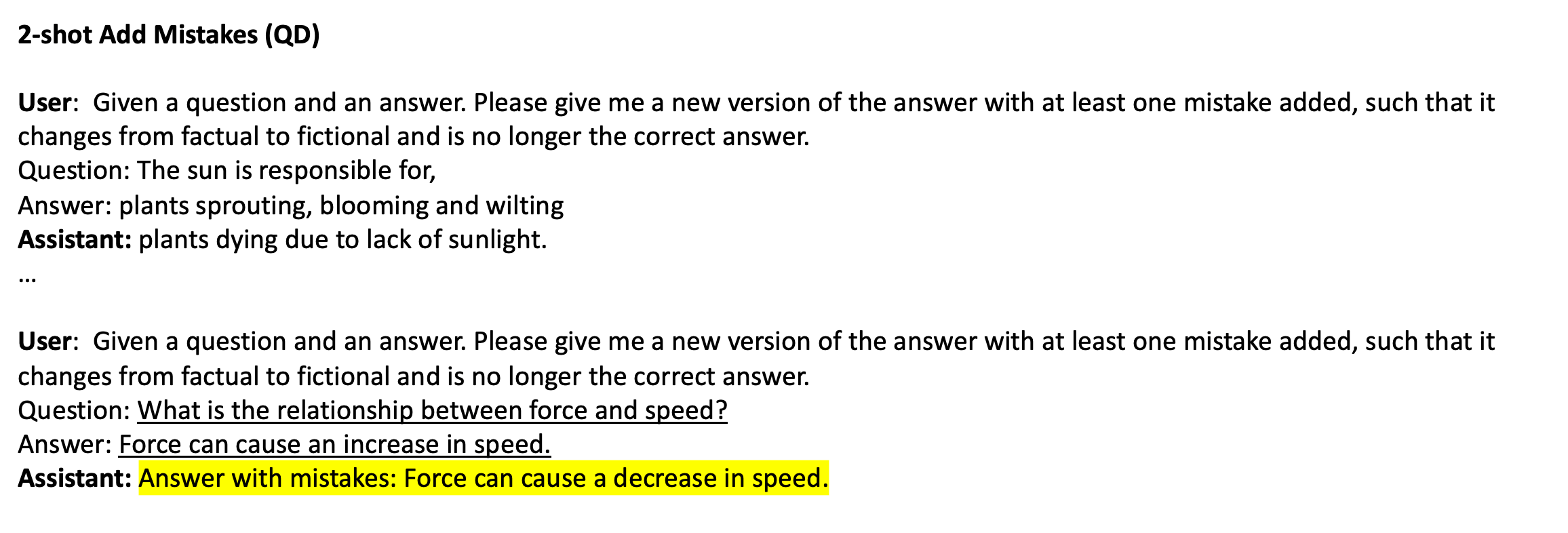}
    \caption{2-shot inserting mistake template for QD. Input [Underline] Generated: [highlighted]. Only show 1 example.} 
    \label{fig:mistake_qd}
\end{figure*}

\begin{figure*}[h]
    \centering 
    \includegraphics[width=1.0\textwidth]{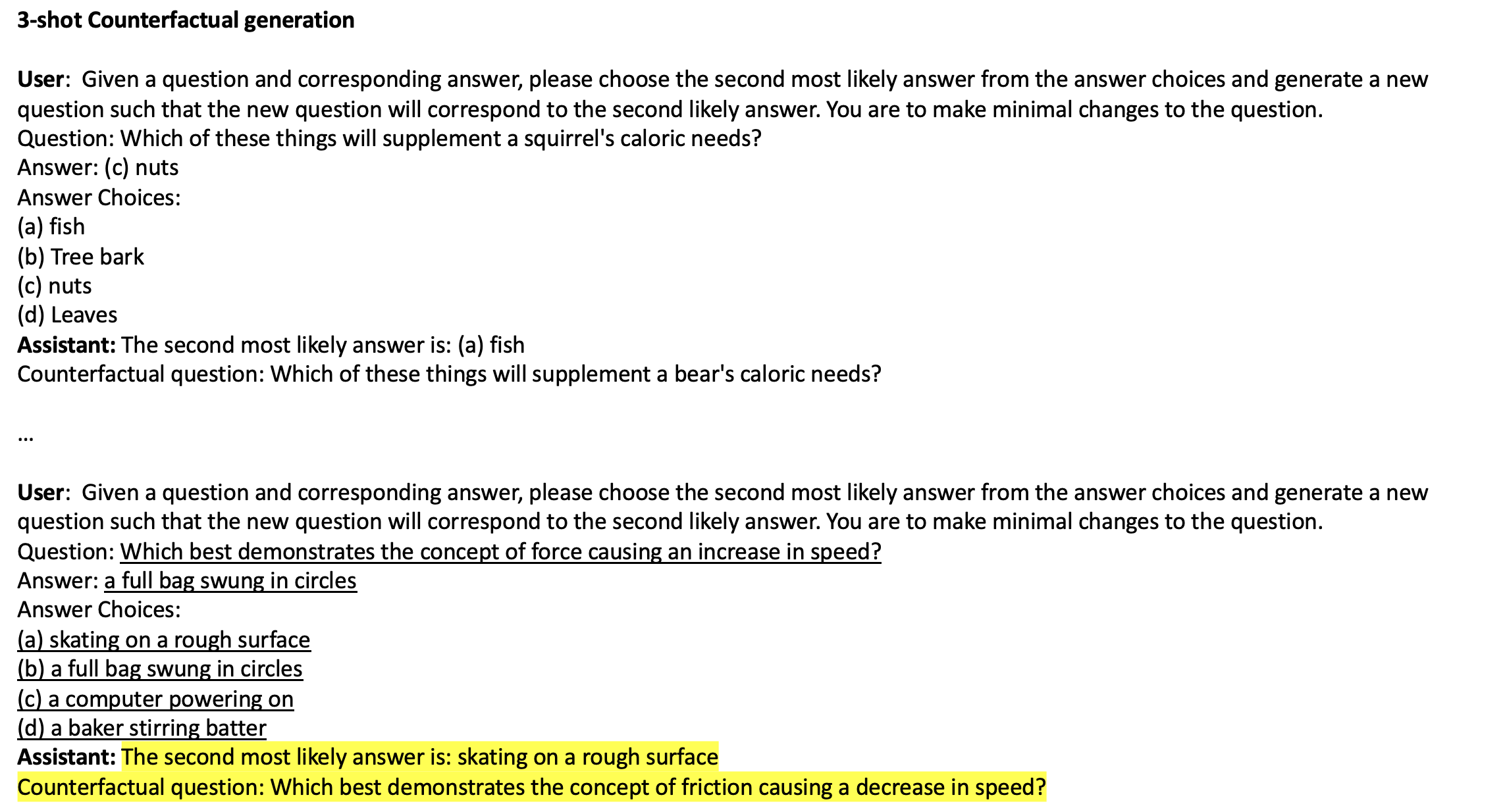}
    \caption{3-shot counterfactual generation Input [Underline] Generated: [highlighted]. Only show 1 example. First, identify the next possible answer before editing the question towards it.} 
    \label{fig:cf}
\end{figure*}

\begin{figure*}[!h]
    \centering 
    \includegraphics[width=1.0\textwidth]{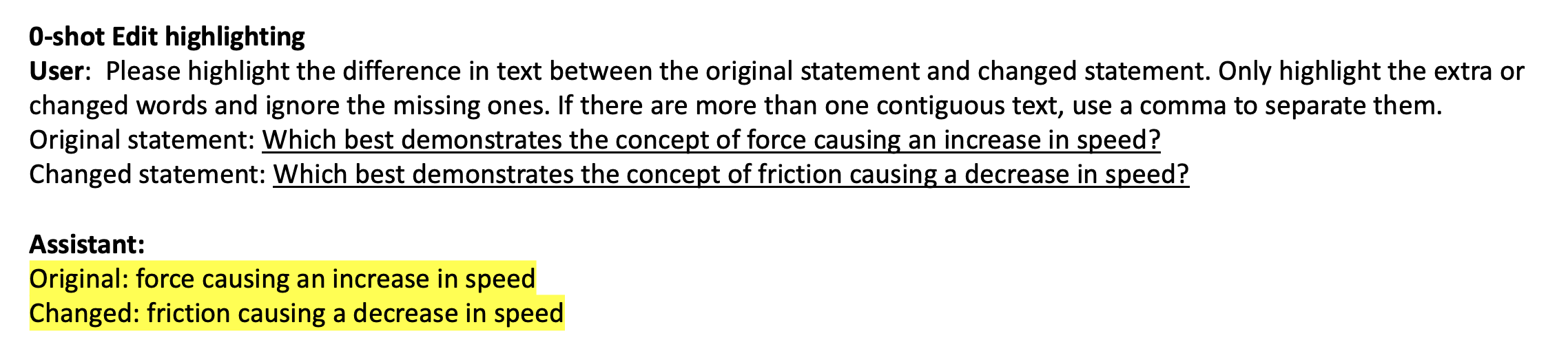}
    \caption{0-shot edit highlighting. Input [Underline] Generated: [highlighted]. Identify edits corresponding to the original text.} 
    \label{fig:cf_edit}
\end{figure*}

We use GPT3.5 to generate paraphrased versions of the reasoning explanation produced by prompting the LLM, except QD. For QD, we select one subquestion-answer pair to apply the perturbations to, we paraphrase both chosen question-answer pairs and only add mistakes to the answer as the focus is on producing wrong answers and not incomprehensible questions. To convert the question $x$ to a counterfactual instance $x'$, we use GPT4 as GPT3.5 frequently produces nonsensical questions that the available answer options cannot answer. Furthermore, we subsequently deploy GPT3.5 again to identify the edited and original portions of $x$, namely the modification $c$. Thus, we end up with two sets of templates for both paraphrasing and addition of mistakes (one for QD, one for others) and one set of counterfactual generation. We use 2-shot examples for adding mistakes, 3-shot for counterfactual generation, and 0-shot for paraphrasing. All figures are from Figure~\ref{fig:para} to~\ref{fig:cf_edit}

\subsection{Inference details}
\label{appendix: inference}
We do not use API for the bulk of the experiments except perturbation generation and ablation using GPT3-5. We mainly rely on local resources to conduct inference. We use 4 x A6000 GPU for all experiments, each GPU has 46GB of VRAM and this gives us a total of 184GB VRAM. A 70B model would require at least 140GB VRAM, leaving only 44 VRAM left for text generation. Given an average input size of 1000 (usually longer for prompts such as QD) and a single batch size of 1, it would require an additional >60 GB VRAM (computed based on L = 80, H= 64, dim = 8192 for 70B) which makes it infeasible to implement. Thus, we perform the experiments using a 4-bit quantized version instead, which is performed using GPTQ on the original Llama-2 70B model. GPTQ is suitable for quantizing models consisting of billions of parameters. It has been validated on models up to 176B parameters and shown comparable performance with 16-bit models. The GPTQ-ed models are readily available on \verb|huggingface|. 

We utilized \verb|text-generation-inference|, an optimized platform for conducting fast inference on LLMs by Huggingface, to speed up the inference process. Overall, this allows us to process up to a batch size of 16 across the full hardware stack.

\subsection{Hyperparameters}
Besides the prompting techniques that use best-of-n preference to select the final output, we stick to greedy decoding. This leaves SC-CoT and SEA-CoT, where we set N to 10 and fix temperature and k to $1.0$ and $50$ respectively while doing sampling. This is only applied during the process of generating explanations, where we revert to greedy decoding during evaluation across all prompting techniques. The number of sequences is set to 10 to balance the computational resources such as RAM and speed. N = 10 is also reported to be sufficient in ~\cite{wang2022self}.

\subsection{Few-shot Prompts}
We show the few-shot examples used for the OBQA dataset, highlighting the differences in the instruction prompt between the various techniques reviewed. The few-shot examples are similar to ~\cite{wei2022chain}, and adjusted when necessary, depending on the specific prompting methodology.

\begin{figure*}[h]
    \centering 
    \includegraphics[width=1.0\textwidth]{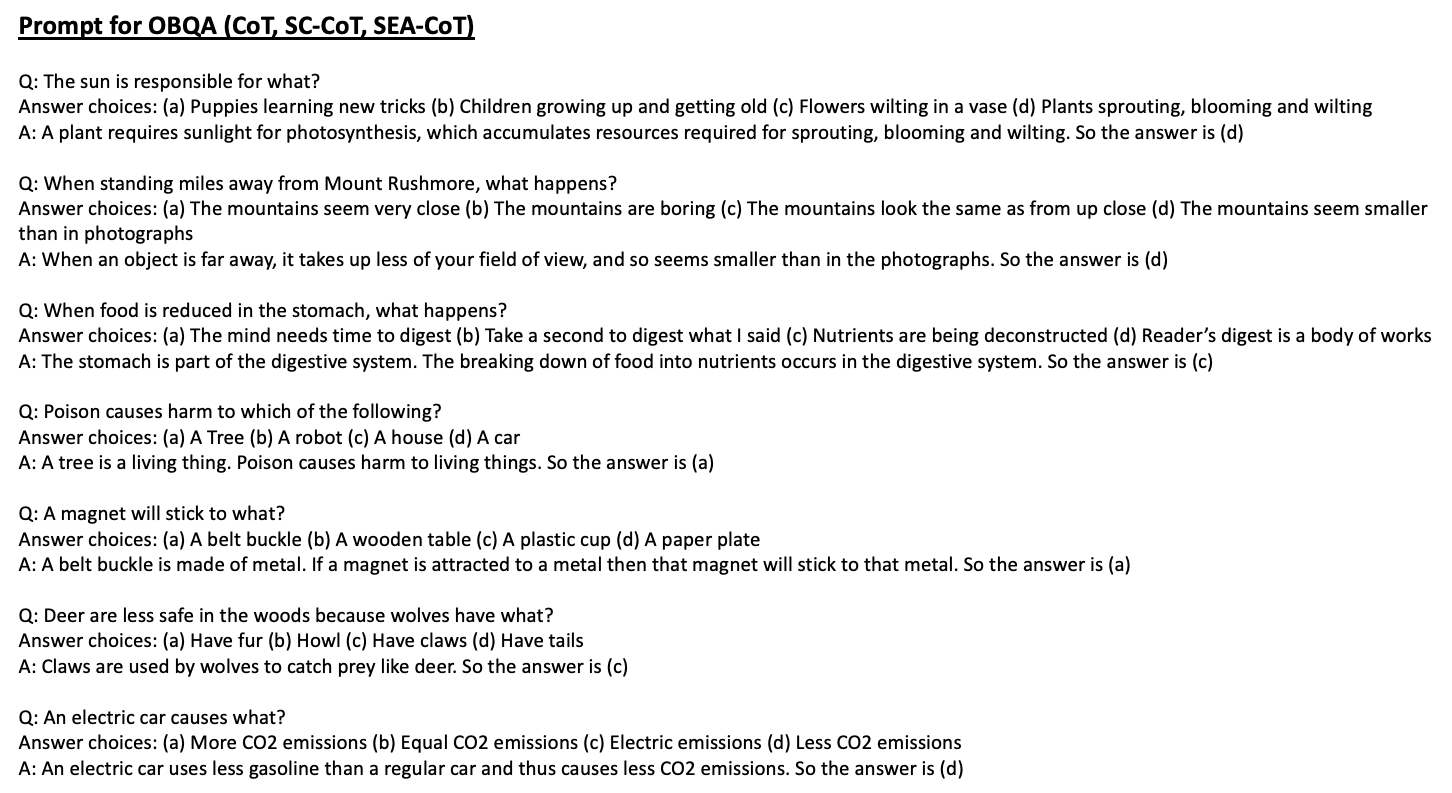}
    \caption{7-shot prompt used for CoT, SC-CoT and SEA-CoT. There are newlines between answer choices and each given choice, is opted out to save space.} 
    \label{fig:obqa_cot}
\end{figure*}

\begin{figure*}[h]
    \centering 
    \includegraphics[width=1.0\textwidth]{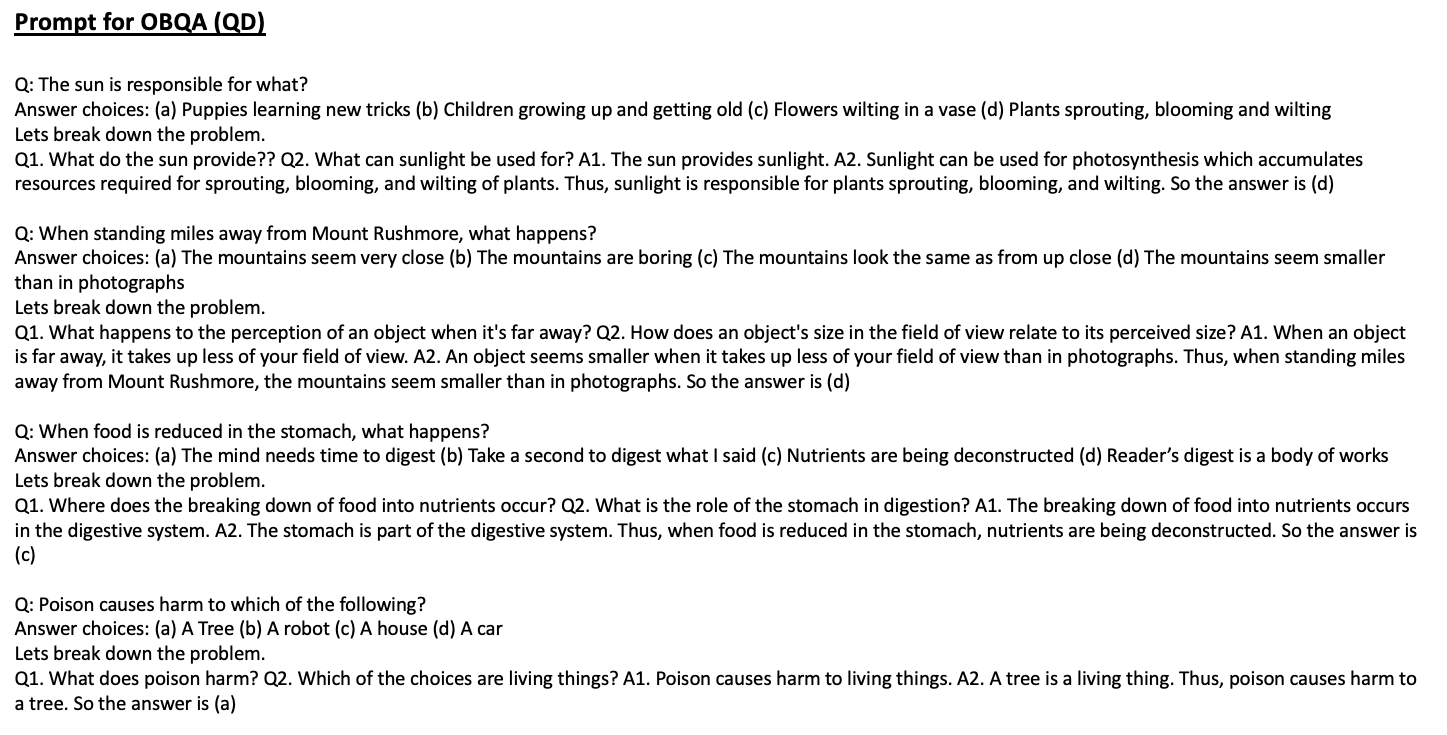}
    \caption{7-shot prompt used for QD. We show only 4 examples here, and there are newlines between each sub-questions and answers, which we similarly leave out to save space.} 
    \label{fig:obqa_qd}
\end{figure*}

\begin{figure*}[h]
    \centering 
    \includegraphics[width=1.0\textwidth]{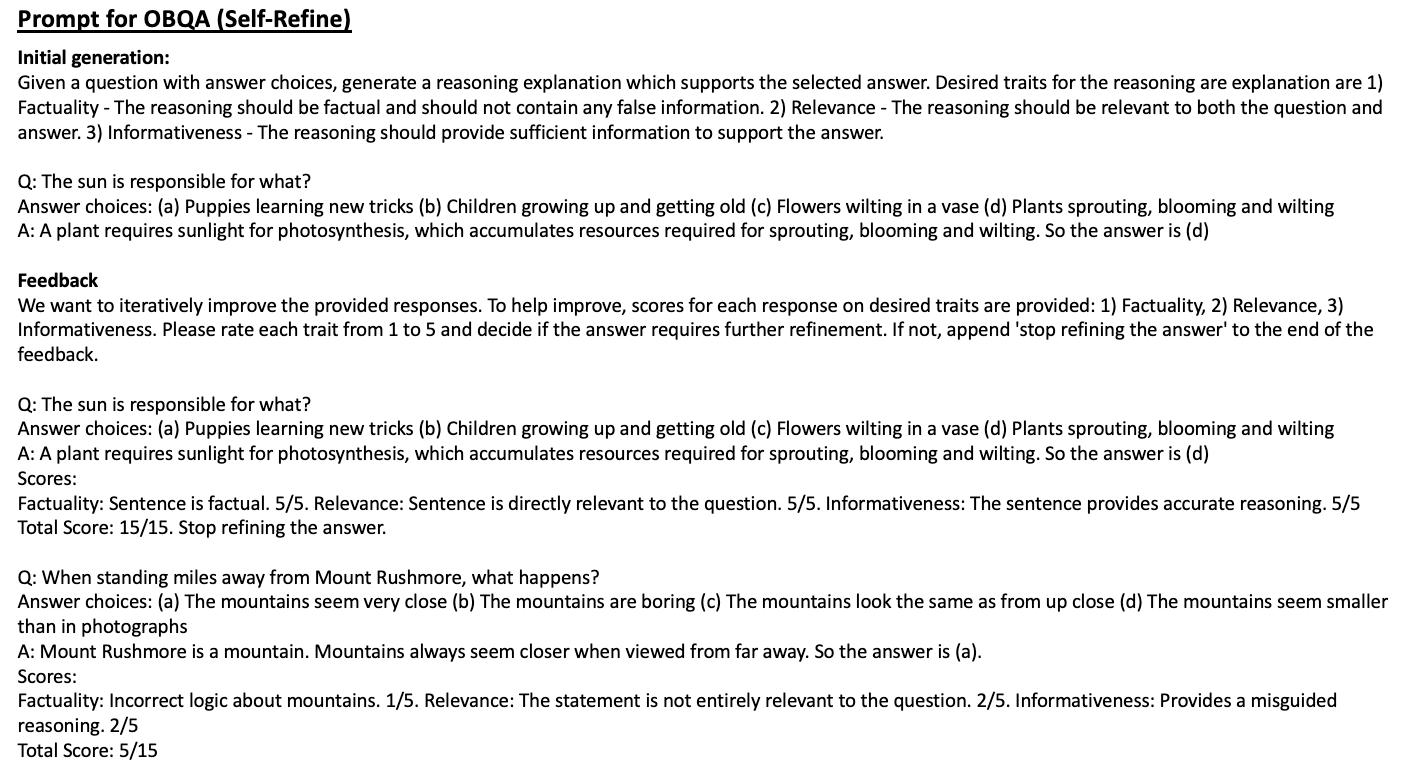}
    \caption{Prompt for Self-Refine, we show a single example for the initial generation, the rest is similar in CoT examples. For the feedback, we include both good and bad examples, both displayed here. We use 7 examples for both initial generation and feedback.} 
    \label{fig:obqa_refine1}
\end{figure*}

\begin{figure*}[h]
    \centering 
    \includegraphics[width=1.0\textwidth]{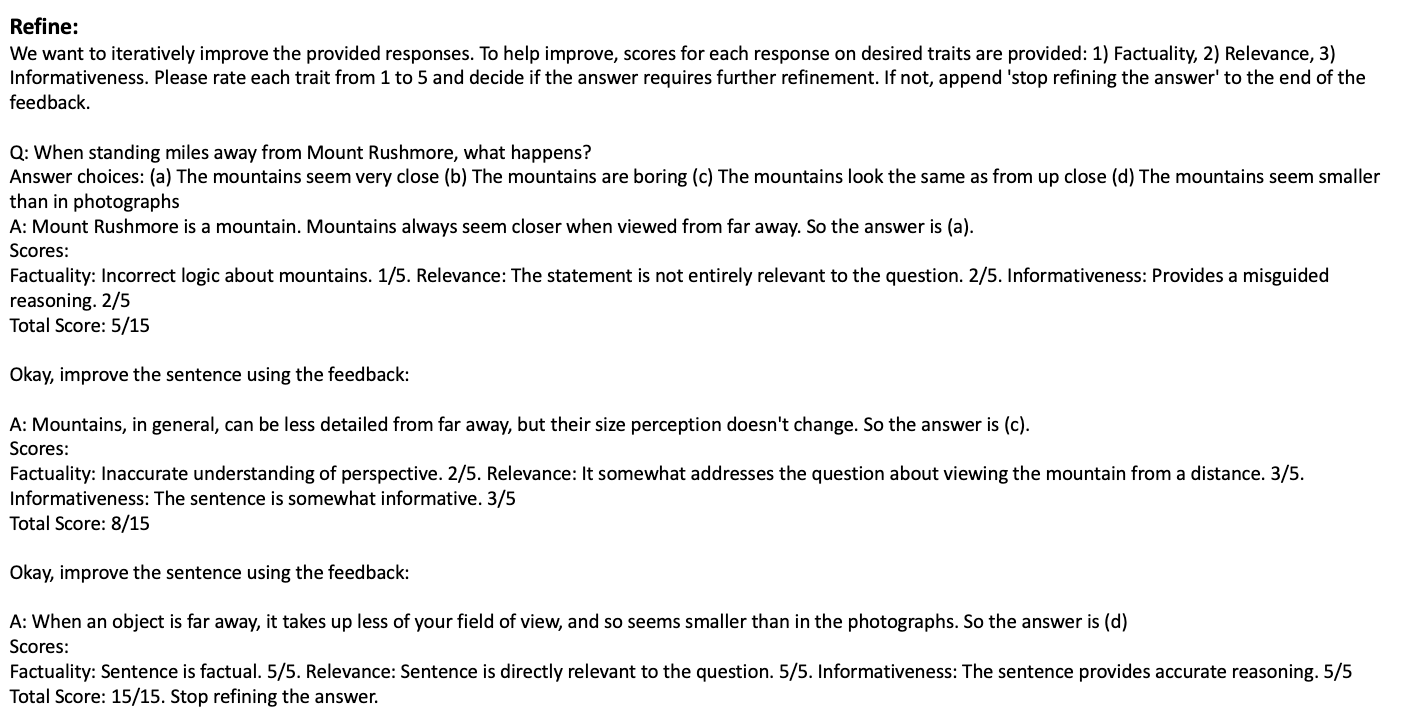}
    \caption{Refine stage in Self-Refine, we show a single example here, where each example demonstrates the entire refining process from a bad to a good example.} 
    \label{fig:obqa_refine2}
\end{figure*}

For Self-Refine, there are three stages of instruction-prompting, where the second (feedback) and third (refine) stages continue iteratively until the LLM detects a stopping criterion which ends the cycle, denoted as \textit{"Stop refining the answer."}. In the \textit{initial generation}, the optimal examples are given, similar to CoT. In the \textit{feedback} stage, we list scoring criteria which are focused on improving the interpretability of the reasoning explanation, instead of focusing on the performance. To simulate various qualities of output, we include both positive and negative examples. The examples in the refine stage are similar to the feedback but are instead designed in a continuous conversion displaying the full process of refining a bad example into a good one. We limit the number of examples in the \textit{refine} stage to 3 as the context length is much longer here. The few-shot example prompts are displayed from Figure~\ref{fig:obqa_cot} to~\ref{fig:obqa_refine2}.

\subsection{Entailment Generation}
We designed a separate prompt to be used solely by SEA-CoT, where the LLM is instructed to self-critique the entailment between its reasoning chain and the combined context of both the question and the produced answer. We use samples from the e-SNLI dataset~\cite{camburu2018snli}, we only picked instances corresponding to either entailment or contradiction and left out the neutral ones, as the LLM is only instructed to infer if the explanation entails or contradicts the target context. 

\begin{figure*}[!ht]
    \centering 
    \includegraphics[width=1.0\textwidth]{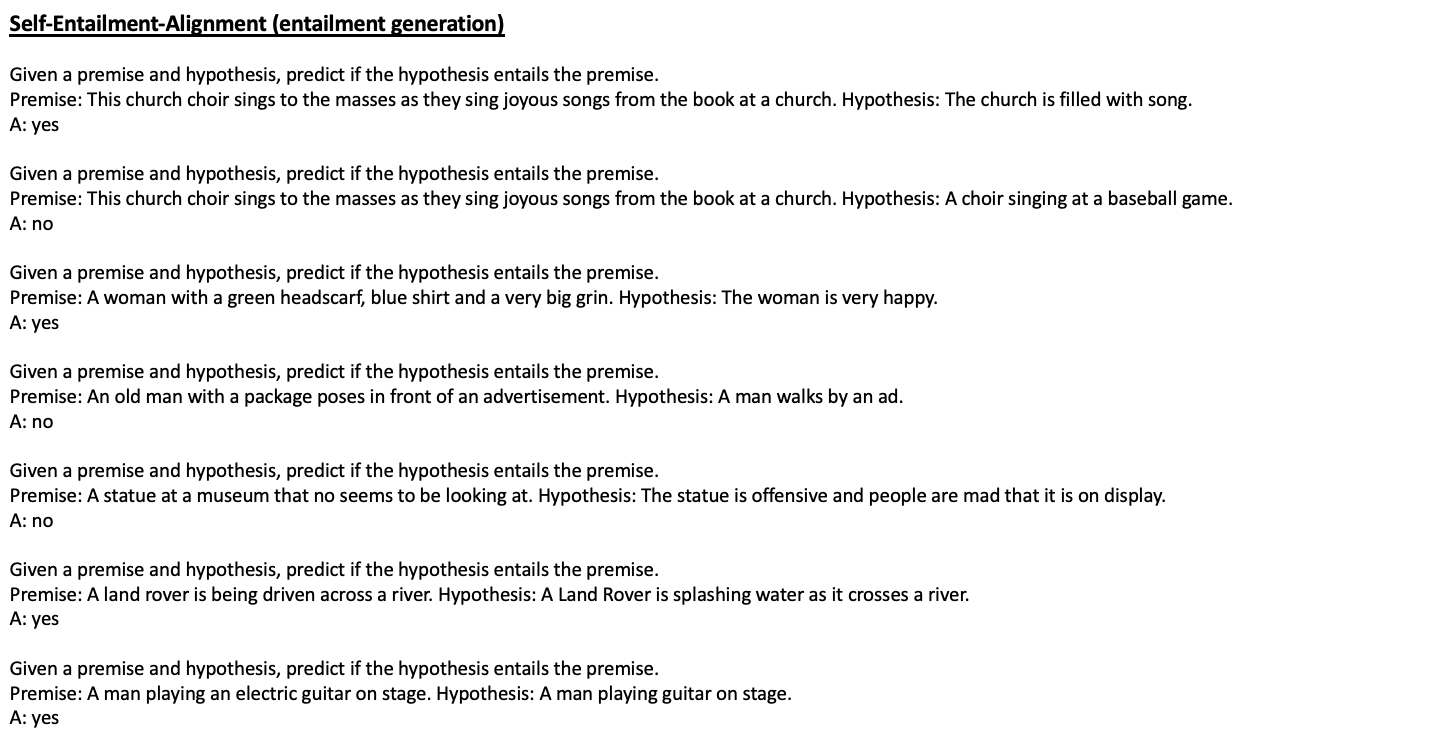}
    \caption{NLI examples for entailment generation for SEA-CoT, used across all datasets.} 
    \label{fig:entailment}
\end{figure*}

The probabilities for the entailment label \textit{``yes''} are directly used while we take the complement if generated \textit{"no"}, with the assumption that other tokens in the vocabulary are negligible. The examples are displayed in Figure~\ref{fig:entailment}.

\subsection{Number of sequences}
We carry out additional experiments on increasing N sequences, to see if increasing the number of options allows the ranking process to select more interpretable explanations. The results in Table~\ref{tab:N values}, showed that increasing N has positive effects on the utility of the reasoning steps, while slight negative effects on the paraphrasing and counterfactual tests. The higher number of sequences may make it difficult to optimize for each quality simultaneously, as one explanation may be more faithful but lacks usefulness in teaching a less technical model to follow its reasoning process. Nonetheless, this study is promising for context distillation, where we may be interested in using the generated response of a larger LLM to teach a smaller model, by using higher N values.

\subsection{Leakge-Adjusted Simulatability (LAS)}
\label{appendix:LAS}
We define the formal definitions of the LAS metric~\cite{hase2020leakage} used in assessing the utility of an explanation here. LAS is primarily used to measure the improvement in task performance upon the addition of an explanation, $\hat{e}$ to the given context, $x$ in producing an outcome, $\hat{y}|x,\hat{e}$. Most importantly, it accounts for the two cases of phenomena encountered. The first is when the model can guess the outcome directly from the input, $x$. In such cases, this renders the explanation, $\hat{e}$ as a false causal input in producing any improvements on the task score. The second is when $\hat{e}$ directly leaks the label to the model and the outcome can be easily guessed without consuming the given question.

The first scenario can be solved by introducing a baseline, $\hat{y}|x$, and subtracting the task performance, $\mathbbm{1}[\hat{y}_i|\hat{e},x_i]$ from $\mathbbm{1}[\hat{y}_i|x_i]$. The second case is accounted for by measuring the performance when the explanation either leaks, $\mathbbm{1}[\hat{y}_i|\hat{e}] = 1$ or not, $\mathbbm{1}[\hat{y}_i|\hat{e}] = 0$. $\mathbbm{1}$ denotes the event where outcome is correctly guessed, $\hat{y} = y$.

The overall LAS score regards both scenarios by taking the average of the subtracted performance in both non-leaking, $LAS_0$, and leaking group, $LAS_1$ below. 
\begin{flalign}
    & LAS_0 = \frac{1}{N_0} \sum_{i=1}^{N_0} (\mathbbm{1}[\hat{y}_i|x_i,\hat{e}_i] - \mathbbm{1}[\hat{y}_i|x_i])\\
    & LAS_1 = \frac{1}{N_1} \sum_{i=1}^{N_1} (\mathbbm{1}[\hat{y}_i|x_i,\hat{e}_i] - \mathbbm{1}[\hat{y}_i|x_i])\\
    & LAS = \frac{1}{2}(LAS_0 + LAS_1)
\end{flalign}
$N_0$ and $N_1$ denote the number of non-leaking and leaking encounters. 

\begin{table}
\begin{tabular}{ccccc}
\toprule
N & P($\downarrow$) & CF-UF ($\downarrow$) & M ($\uparrow$) & S ($\uparrow$)\\
\midrule
10 & \textbf{1.2} & \textbf{3.81} & 61.24 & 16.97 \\
30 & 2.01 & 5.98 & 67.77 & 17.2 \\
50 & 1.8 & 6.49 & \textbf{68.40} & \textbf{18.7} \\
\bottomrule
\end{tabular}
\caption{Interpretability scores across different numbers of sequences generated per sample.}
\label{tab:N values}
\end{table}



\end{document}